\documentclass[11pt,a4paper]{article}
\usepackage[T1]{fontenc}
\usepackage[utf8]{inputenc}
\usepackage[a4paper,left=23mm,right=23mm,top=25mm,bottom=23mm]{geometry}
\usepackage{amsmath,amssymb,amsthm}
\usepackage{newtxtext,newtxmath,microtype}
\usepackage{graphicx,xcolor,booktabs,array,tabularx,longtable}
\usepackage{enumitem,titlesec}
\usepackage{placeins,float}
\usepackage{chessboard}
\setsidefontshape{n}
\usepackage[font=small,labelfont=bf]{caption}
\usepackage[round,authoryear]{natbib}
\usepackage{xurl}
\usepackage[hidelinks]{hyperref}
\hypersetup{pdftitle={RetroChessBench: A Benchmark for Reconstructing and Proving Chess Histories},pdfauthor={Pricope Tidor-Vlad, D'Arcy Jocelyn Kate},pdfsubject={Retrograde chess, language-model evaluation, and terminal reward environments}}
\usepackage{bookmark}
\newcolumntype{Y}{>{\raggedright\arraybackslash}X}
\newcommand{\bench}{RetroChessBench}
\newcommand{\model}{GPT 6 Astra High}
\newcommand{\sq}[1]{\mathrm{#1}}
\newcommand{\code}[1]{\texttt{#1}}

\newcommand{\dataseturl}{https://app.poindexterlabs.net/docs/retro-chess}
\newcommand{\benchmarkurl}{\dataseturl?tab=mini}
\newcommand{\collectionurl}{\dataseturl?tab=all}
\newcommand{\environmenturl}{\dataseturl?tab=mini\#environment}
\newcommand{\environmentzipurl}{\dataseturl/downloads/poindexter-retro-rl-2026-10-27.zip}
\newcommand{\environmentsetupurl}{\dataseturl/downloads/README.txt}
\newcommand{\environmentvalidationurl}{\dataseturl/downloads/VALIDATION.txt}
\newcommand{\huggingfaceurl}{https://huggingface.co/datasets/Poindexter-Labs/PDL-Bench}
\newcommand{\hfenvironmenturl}{\huggingfaceurl/blob/main/environments/retro-chess/README.md}
\newcommand{\hfzipurl}{\huggingfaceurl/resolve/main/environments/retro-chess/poindexter-retro-rl-2026-10-27.zip}
\newcommand{\gradingurl}{\dataseturl/grading/index.html}

\setlist{topsep=4pt,itemsep=2pt,parsep=0pt,leftmargin=1.5em}
\titleformat{\section}{\large\bfseries}{\thesection}{.65em}{}
\titleformat{\subsection}{\normalsize\bfseries}{\thesubsection}{.65em}{}
\titlespacing*{\section}{0pt}{13pt}{5pt}
\titlespacing*{\subsection}{0pt}{9pt}{3pt}
\begin{document}

\begin{center}
\vspace*{2mm}
{\LARGE\bfseries RetroChessBench: A Benchmark for\\[3pt]
Reconstructing and Proving Chess Histories\par}
\vspace{4mm}
{\large Pricope Tidor-Vlad}\par
Poindexter Labs\par
\vspace{2mm}
{\small 27 October 2026}
\end{center}

\begin{abstract}
Frontier language models such as GPT 6 Astra have contributed to research mathematics, demonstrating substantial reasoning capabilities. Yet serious reasoning failures persist even in tasks governed by explicit rules. We demonstrate this in retrograde chess which is a playground for pure reasoning, where solvers reconstruct legal histories and distinguish possible events from forced ones. We introduce \bench{}, a 12-task benchmark covering exact-length constructions, unique endings, and exhaustive answer sets, with references, rubrics, and a terminal-reward environment. In two trials per task and condition, \model{} fails to produce the required final answer in 22/24 attempts without tools and 18/24 with code execution; mean terminal RL rewards are 14.75\% and 27.19\%, respectively. The examples identify faulty deductions: failed responses reject legal histories and treat temporary obstructions as permanent. Online retrograde archives often provide puzzles and final move sequences without detailed derivations. We hypothesize that limited exposure to these missing explanations in web-derived training data contributes to the observed reasoning failures. This questions how far frontier mathematical reasoning really generalizes beyond familiar solution patterns. Claims of general intelligence require evidence that models can transfer learnt principles to areas with few worked examples.
\end{abstract}

\section{Introduction}
A chess position records where the surviving pieces stand. It omits the moves that brought them there, the pieces that disappeared, and the promotions that may have occurred. Retrograde analysis reconstructs this missing history under the ordinary rules of chess. The scientific question is whether a model can reason consistently over these hidden histories and distinguish what \emph{could} have happened from what \emph{must} have happened.

The distinction changes the standard of proof. An exact-length proof game needs one legal witness. A unique-ending problem needs an argument covering every compatible history. A problem with several possible last moves needs a feasible history for each answer and an exclusion of all others. These obligations remain different even when their final answers fit on one line. A six-move ending can depend on a promotion or capture budget established much earlier in the game.

These tasks put a basic expectation of general reasoning under pressure: known rules should remain usable when the question changes from choosing a future move to establishing a necessary past event. The solver must keep track of hypothetical states, revise assumptions, and distinguish constraints that hold in one candidate history from constraints that hold in all of them. Fluent reconstruction can conceal a failure at precisely this distinction.

A concrete failure illustrates the issue. In P0292, a model constructs a legal 47-ply game and concludes that the requested 45-ply game is impossible. Its bound treats a pawn on g6 as an unavoidable obstruction to a bishop's route. A replayable 45-ply construction sends the bishop through g6 \emph{before} the pawn arrives. The model has found one possible history and mistaken its schedule for a necessary constraint. Table~\ref{tab:openingcase} isolates the error; Appendix~\ref{app:witness} gives the full counterexample.

We introduce \href{\benchmarkurl}{\bench{}} (also on \href{\huggingfaceurl}{Hugging Face}), a 12-task benchmark comprising seven exact-length games, four unique suffixes, and one exhaustive two-answer task. These tasks were selected from a \href{\collectionurl}{48-task candidate pool} drawn from \emph{StrateGems} compositions, including two adaptations. References, prompts, board images, responses, rubrics, and replay checks accompany the tasks. A \href{\environmenturl}{terminal-reward interface} accepts alternative valid constructions and scores analytical obligations separately.

We evaluate \model{} twice per task without tools and twice with hosted code execution. Final answers are correct in 2/24 and 6/24 attempts. The code condition produces two valid constructive games and four correct suffixes; three of those suffix answers still lack complete proofs. Five construction tasks fail in all four attempts. These are descriptive outcomes on a set selected using observed failures, not an unbiased estimate of retrograde-chess ability or a causal estimate of tool benefit.

OpenAI reported a Navier--Stokes solution on 8 September 2026 \citep{openai2026navier}. Four weeks later, on 6 October, it released hundreds of mathematical research manuscripts with varying verification status \citep{openai2026mathrelease,openai2026math}. Both releases concern internal models; OpenAI describes the Navier--Stokes model as more capable than Astra, whose separate mathematical contributions include prime-gap results \citep{openai2026astra}. ChessBench's 27 October 2026 snapshot lists Astra at 1745 Elo for XHigh and 1531 for High \citep{chessbench2026astra}. These reports motivate broader tests of reasoning. The question is whether mathematical progress brings reliable construction and exclusion across other domains.

Backward reasoning is already studied in language-model planning, and reverse chess appears in LongCoT \citep{ren2024backward,motwani2026longcot}. Our source compositions and answers are public, but availability does not establish training exposure. A leading hypothesis is limited exposure to worked retrograde explanations. The source archive gives P0292's game with a short theme description and R0185's ending with a compact rejected branch \citep{jankoarchive}. A game certifies existence but need not teach discovery; a suffix sequence alone does not establish uniqueness. Detailed treatments do exist \citep{volet2003inference}; the hypothesis concerns explanatory coverage in training, which we have not measured. If such coverage is sparse, can a model assemble familiar rules into historical deductions without abundant demonstrations of the method?

Our contribution operationalizes these established proof obligations in hidden chess histories through distinct answer contracts, executable witness checks, recorded proof assessments, and inspectable counterexamples to failed arguments. The benchmark and auditable pilot motivate controlled tests of local rule knowledge, candidate discovery, state representation, and proof verification (Section~\ref{sec:discussion}). Those tests can distinguish missing techniques from failures to coordinate techniques the model already possesses.

% Keep this worked example after the complete introduction; do not float it.
\begin{table}[H]
\centering\small
\begin{tabularx}{\linewidth}{@{}lY@{}}
\toprule
P0292, code run 1 & Observable argument and counterexample\\
\midrule
Invalid inference & A pawn on g6 prevents Be8-h5, so the required construction needs at least 47 plies.\\[2pt]
Legal counterexample & The 45-ply reference contains 6.e8=B+ Ke6 7.Bh5 g6 8.g5 gxh5: the bishop traverses g6 before it is occupied.\\[2pt]
Missing obligation & A lower bound must cover alternative event orders. A valid 47-ply witness establishes no such bound.\\
\bottomrule
\end{tabularx}
\caption{A failure of historical exclusion. The claim is paraphrased from the saved final answer; the counterexample is independently replayed from the initial position. This example establishes an error in the argument, not its internal computational cause.}\label{tab:openingcase}
\end{table}

\section{Related work}
\paragraph{Chess play and state tracking.}
Chess offers several distinct evaluation targets. \citet{toshniwal2022state} study piece tracking and legal-move prediction from move sequences, finding that explicit state information helps with limited training data. \citet{harang2025tracking} evaluate inferred chess states through the actions those states permit. \citet{kolasani2025chess} examine reasoning and instruction following during extended forward play. ChessBench derives ratings from game outcomes using an anchored Monte Carlo Elo procedure \citep{chessbench2026methods}. These settings test following or choosing a continuation from a supplied history. \bench{} instead asks the solver to recover unobserved events, including captures and promotions, and sometimes to prove that other histories cannot change the answer. A forward-play rating motivates comparison but supplies no calibration for historical proof difficulty.

\paragraph{Reverse reconstruction and classical planning.}
LongCoT is the closest benchmark predecessor: its reverse-chess task reconstructs a legal alternating move sequence consistent with a terminal board, side to move, and specified move fragments \citep{motwani2026longcot}. This shares the existential objective of our proof games. The unique-suffix and exhaustive-answer tasks add universal exclusions over legal histories. PlanBench provides systematic tests of planning and reasoning about actions and change using domains from automated planning \citep{valmeekam2023planbench}. It establishes a broader methodological precedent for explicit transition rules and checkable plans. \citet{ren2024backward} study directional planning biases and combine forward and backward search with verification. Together these studies motivate separating discovery from checking, rather than treating all backward reasoning as one capability.

\paragraph{Abduction and missing premises.}
Recovering history is a form of reasoning from effects to possible causes. ART asks models to choose or generate plausible explanations of narrative observations \citep{bhagavatula2020abduction}. Plausibility is a different criterion from chess legality, and neither entails uniqueness. ProofWriter generates implications, proofs, and abductive statements over explicit natural-language theories \citep{tafjord2021proofwriter}. \citet{sprague2022incomplete} interleave forward deduction and backward abduction to infer missing premises, using round-trip validation to filter candidates. \bench{} shares the need to propose hidden events and check their consequences, but the missing objects are time-indexed states and piece identities under a fixed transition system. An explanation must fit a complete legal history; a uniqueness claim must survive every alternative explanation admitted by that system.

\paragraph{Proof search, quantifiers, and generalization.}
PrOntoQA makes generated reasoning steps amenable to formal analysis. \citet{saparov2023greedy} find that valid individual deductions can coexist with failures to explore alternative proof steps. \citet{saparov2023ood} separately test depth, width, and compositional generalization, including hypothetical subproofs. These results motivate our distinction between knowing a local chess rule and coordinating it across a branching historical argument; they do not establish the cause of Astra's errors. SATBench evaluates constraint satisfaction and unsatisfiability on generated logical puzzles \citep{wei2025satbench}, a related separation of witnesses from exclusion. \bench{} adds capture identities, alternating turns, and hidden event order, while retaining the need to distinguish an existential solution from a universal claim.

\paragraph{Mathematics, constructions, and robustness.}
MathArena evaluates recent competition mathematics, and Proof or Bluff examines proof quality on olympiad problems \citep{balunovic2025matharena,petrov2025proof}. MathConstruct accepts automatically checkable mathematical constructions \citep{balunovic2025mathconstruct}; the same principle governs our exact-length games: any valid witness earns full credit. Those results provide context for asking about transfer, rather than a measurement of transfer in this study. GSM-Symbolic probes mathematical robustness through template variations \citep{mirzadeh2025symbolic}. Its later re-evaluation identifies numerical-distribution and statistical confounds and reports model-specific failure profiles \citep{dlugosz2026earnest}. The methodological lesson is relevant here: a performance gap requires controlled explanation. Counterfactual-task evaluations \citep{wu2024counterfactual} and chess-prior interventions \citep{pleiss2026generalization} offer complementary ways to test dependence on familiar representations. \bench{} keeps orthodox rules fixed; future transfer tests must additionally control the inference demanded and the solution budget.

\paragraph{Classical retrograde analysis and assessment.}
The compositions and source solutions predate this benchmark and remain credited to their authors \citep{jankoarchive}. Generalized retrograde reachability on growing boards is PSPACE-complete \citep{brunner2020retro}; this does not classify the difficulty of our fixed $8\times8$ instances. Specialist software such as Jacobi already solves proof games \citep{labellejacobi}. Our claim concerns language-model evaluation, not the impossibility of algorithmic reconstruction. Natural-language uniqueness proofs introduce a different assessment problem. Known limitations of model judges \citep{zheng2023judge} require separate records for deterministic witness checks, rubric credit, and proof review. A reference receiving full credit is a calibration control, not independent evidence that the proof or judge is correct.

\begin{table}[htbp]
\centering\small
\begin{tabularx}{\linewidth}{@{}>{\raggedright\arraybackslash}p{.22\linewidth}YY@{}}
\toprule
Evaluation setting & Requested object & Relevant distinction\\
\midrule
Forward chess & Legal or strong next moves; completed games & Continuation quality does not establish a hidden history.\\[2pt]
LongCoT reverse chess & A history satisfying terminal and sequence constraints & A construction establishes existence.\\[2pt]
PlanBench & Plans and reasoning about state changes & Explicit action models permit separate planning and checking tasks.\\[2pt]
ART / ProofWriter & Plausible missing causes / deductions and abductive premises & Plausibility, derivability, and necessity are different objectives.\\[2pt]
Formal deduction / satisfiability (\mbox{PrOntoQA}; \mbox{SATBench}) & Formal derivations; satisfiability and unsatisfiability decisions & Proof evaluation and excluding all satisfying assignments are established objectives.\\[2pt]
\bench{} & Exact-length games, unique suffixes, complete ending sets & Distinct answer contracts apply these obligations to hidden histories, with replayable witnesses and inspectable failures.\\
\bottomrule
\end{tabularx}
\caption{Relationship to prior evaluation settings. This comparison identifies the task objective; it does not rank benchmarks or claim that historical reconstruction is new. The cited implementations differ in models, budgets, and task distributions.}\label{tab:related}
\end{table}

\section{Tasks and proof obligations}\label{sec:tasks}
Let $s_0$ be the standard initial state, $T(s,m)$ the legal transition for move $m$, and $D$ a terminal piece placement. A full state also includes side to move, castling rights, and en-passant rights. The supplied FEN \emph{piece-placement field} leaves these unspecified unless the task determines them. Write $\pi(s)$ for a state's placement. The set $\mathcal H_C(D)$ contains histories $h=(m_1,\ldots,m_n)$ that start at $s_0$, follow legal transitions, end with $\pi(s_n)=D$, and satisfy the task conditions $C$.

For answer comparison, each transition induces an annotated event $e_i=(o_i,d_i,p_i,v_i)$: origin, destination, promotion type (or none), and captured-piece type (or none). Write $E(h)=(e_1,\ldots,e_n)$. The capture type is evaluated in the predecessor state; en-passant captures also use that state's rights. Check and mate markers are derived from the resulting state and are not part of $E(h)$; Section~\ref{sec:grading} specifies their separate treatment in binary outcomes and rewards. This representation matters: b7$\times$Rc8=R\# and b7$\times$Nc8=R\# both encode as UCI \code{b7c8r}, but are different answers to R0171. A comparison that discards the victim merges two required alternatives.

\subsection{Exact-length constructive proof games}
Given $D$ and $n$, produce any $h\in\mathcal H_{|h|=n}(D)$. Replay checks every move, the number of plies (single moves), and the final placement. Both sides cooperate to reach the diagram. Any valid sequence proves existence and satisfies the task. Uniqueness and minimum length are separate claims.

\subsection{Unique suffixes}
Given $D$ and $k$, assuming legal reachability as stated in the prompt, determine a sequence $\sigma$ such that
\begin{equation}\label{eq:suffix}
  \operatorname{suffix}_k(E(h))=\sigma
  \qquad\text{for every }h\in\mathcal H_C(D).
\end{equation}
Many complete histories may share the same final $k$ plies. Equation~\eqref{eq:suffix} requires excluding all competing endings, including locally legal predecessors that have no legal earlier history. Replaying one candidate ending cannot establish this claim. A benchmark also needs $\mathcal H_C(D)\neq\varnothing$: otherwise the universal statement is vacuous. The prompts explicitly let solvers assume reachability, but that does not discharge the benchmark author's existence obligation. The four suffix references in this bundle provide conditional analytical arguments and local suffix checks, not complete start-to-diagram games. Supplying those witnesses, particularly for R01's changed rook placement, is an outstanding validation step. Appendix~\ref{app:audit} distinguishes this gap from the fully replayed construction and R0171 controls.

\subsection{Exhaustive alternative endings}
R0171 asks for both possible last moves. Its target is a set,
\begin{equation}
  A=\{\operatorname{suffix}_1(E(h)):h\in\mathcal H_C(D)\},
\end{equation}
with two elements. Full credit requires both alternatives, a feasible history for each, and exclusion of further possibilities.

\subsection{Why a standard chess engine is insufficient}
Stockfish searches forward for strong continuations \citep{stockfishdocs}. Historical reconstruction requires additional information: the identity of captured pieces, earlier promotions, and the times at which squares were available. A best-move query does not supply it. Undoing a recorded move also differs from generating every possible predecessor of an arbitrary board.

Code can still help. \code{python-chess} provides move generation, legality checks, notation handling, and replay \citep{fiekaschess}. Custom search, constraint solvers, and proof-game programs can reconstruct histories. Our code condition supplies general hosted execution; it does not supply a complete retrograde solver. Historical reconstruction also differs from computing game-theoretic endgame tablebases backward from terminal states.

\section{Benchmark construction}\label{sec:data}
\subsection{From the source archive to the benchmark}
We audited 82 primary problem entries in \emph{StrateGems} Nos.~52--58 and documented their variants. Eligible tasks asked for prior moves or an orthodox proof game. We excluded piece-addition tasks, position release alone, and fairy-chess stipulations. The resulting 58 task instances each received two no-code runs. At least one run failed on 47 instances. We removed P0299 because its analytical completeness proof remained unresolved. The remaining 46 instances and two adaptations form a 48-task candidate pool.

\bench{} contains 12 tasks with completed grading and recorded secondary review of all 48 solver responses. Ten were selected using observed results, with R0178 and the easier R01 adaptation included to broaden suffix coverage. The final inventory and attempt slots are frozen, but the selection process was adaptive: it first retained no-code failures and subsequently used code outcomes and proof coverage to select the benchmark tasks. The reported no-code trials include the same attempts used for screening. This creates an asymmetric comparison: screening for no-code failure can make subsequent code trials look better through selection and regression to the mean, even without a tool effect. Later selection on code failures adds another dependency. The retained snapshots document the filtering stages and final task IDs, not a preregistered selection rule. We report a challenge-set baseline and give the selection ledger in Appendix~\ref{app:inventory}. A causal tool comparison requires a frozen task set and fresh trials in \emph{both} conditions.

\subsection{Composition, adaptations, and attribution}
The benchmark comprises seven constructive tasks, four unique-suffix tasks, and one two-alternative task. Constructive lengths range from 35 to 60 plies; suffix lengths are 5, 6, 36, and 40 plies. Table~\ref{tab:inventory} credits every composer and source issue.

R01 adapts Andrey Frolkin's R0185 (\emph{StrateGems} No.~56, 2011). We ask for the last six single moves instead of eight because this shorter ending retains the central reasoning challenge; proving its uniqueness still requires looking further back. We also relocate White's rook from h8 to g8, leaving every other piece in place. The rook remains to the right of Qd8, preserving its role in the eighth-rank obstruction used by the proof. The relocation deliberately changes the original diagram to reduce reliance on exact diagram memorization, should the source have appeared in training. It does not establish absence from pretraining data or measure a reduction in memorization.

R02 adapts Nikolai I. Beluhov's R0176 (\emph{StrateGems} No.~53, 2011). The diagram is unchanged; the only task modification is to request the last five single moves instead of forty. This keeps the exposition manageable while preserving the central trap, which arises near the start of the backward analysis. Establishing those five moves still requires excluding a tempting continuation whose failure becomes apparent only further back in the history.

Both adaptations retain the central proof mechanism under the prompt's legal-reachability premise. R02 and R0176 share a position and much of their reasoning; Section~\ref{sec:results} reports a family-weighted sensitivity analysis.

\begin{figure}[t]
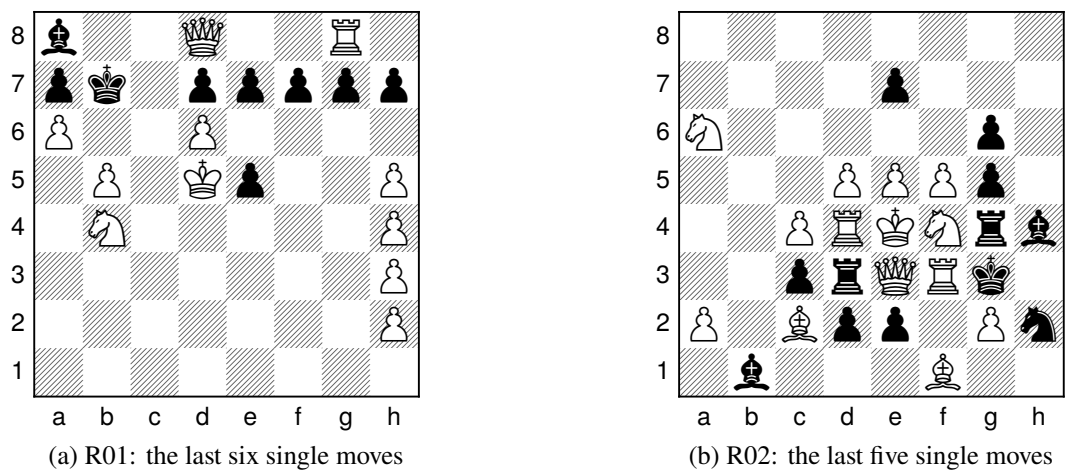

\centering
\begin{minipage}[t]{.48\textwidth}\centering
\chessboard[setfen={b2Q2R1/pk1ppppp/P2P4/1P1Kp2P/1N5P/7P/7P/8 b - - 0 1},boardfontsize=18pt,showmover=false]
\par\small (a) R01: the last six single moves
\end{minipage}\hfill
\begin{minipage}[t]{.48\textwidth}\centering
\chessboard[setfen={8/4p3/N5p1/3PPPp1/2PRKNrb/2prQRk1/P1Bpp1Pn/1b3B2 b - - 0 1},boardfontsize=18pt,showmover=false]
\par\small (b) R02: the last five single moves
\end{minipage}
\caption{Two adapted short-answer tasks. R01 changes Frolkin's R0185 from eight to six single moves and relocates White's rook from h8 to g8. R02 changes Beluhov's R0176 from forty to five single moves, with no change to the diagram. Both retain the central historical proof challenge and assume legal reachability. The solver receives both the diagram and its FEN placement.}\label{fig:boards}
\end{figure}

\subsection{What the artifact contains}
Tasks include prompts, FENs, board images, attribution, reference proofs, answer specifications, and versioned rubrics. Evaluation records contain responses, request metadata, criterion judgments, sequence checks, reviews, and hashes. Each solver rollout is an episode; its trajectory receives a terminal reward and feedback based on the final answer and justification. Intermediate messages and tool calls receive no separate process rewards. This supports reinforcement learning; we evaluate without training a policy.

The \href{\benchmarkurl}{benchmark website} provides task documents, paired results, and evidence; \href{\huggingfaceurl}{PDL-Bench on Hugging Face} hosts the 12-task \code{retro\_chess} subset. The \href{\environmenturl}{RL environment} is available as a \href{\environmentzipurl}{tested Docker build bundle}, also on \href{\hfenvironmenturl}{Hugging Face}, with \href{\environmentsetupurl}{setup instructions} and a \href{\environmentvalidationurl}{validation report}. Researchers can export solver inputs, check references, and grade new trajectories against the frozen tasks and rubrics. The paper bundle includes scripts that reproduce the reported tables and aggregate scores from archived judgments. Appendices~\ref{app:protocol} and~\ref{app:inventory} specify the protocol and archived release.

References and rubrics were developed with AI assistance. The author manually reviewed the solutions on a physical chessboard. A Codex agent performed the recorded secondary reviews. The artifact records these roles and retains attribution for the original compositions and solutions.

\section{Reward design and grading}\label{sec:grading}
\subsection{Separate answer correctness from proof quality}
The evaluator reports final-answer correctness, categorical proof status, and terminal reward separately. A final-answer failure means the required game, suffix, or complete answer set was not produced. Terminal reward scores the task's full objective. A correct ending with an incomplete proof can earn less than 100\%; sound deductions can earn partial credit despite a wrong ending. The public dashboard calls a final-answer failure a ``stump.'' The primary scientific outcomes are exact-answer success and whether the applicable proof obligation is complete. Aggregate reward is an environment-specific summary, not a percentage of reasoning ability or of tasks solved.

The binary answer metric uses verified sequence status, not full sequence-criterion credit. Omitting or mislabeling \code{+} or \code{\#} does not turn an otherwise verified game, suffix, or complete answer set into a final-answer failure. Such defects can reduce analytical sequence credit and hence reward; they do not reduce the binary constructive witness reward. Required move order, promotion identities, and captured-piece alternatives remain substantive constraints. All 48 retained verdicts agree with this rule, and none of the 40 failures is notation-only (Appendix~\ref{app:audit}).

\subsection{Deterministic witness rewards}
For a constructive task and submitted candidate $h$, the reward is
\begin{equation}\label{eq:witness}
r_{\mathrm{game}}(h)=
\begin{cases}
1,&h\text{ replays legally from }s_0,\ |h|=n,\ \pi(s_n)=D,\\
0,&\text{no valid complete witness is supplied.}
\end{cases}
\end{equation}
The checker accepts equivalent move encodings and checks supplied piece and capture identities. It reports check-annotation errors separately from move legality. A valid witness earns full reward regardless of its agreement with the reference or the length of its explanation. A legal game that arrives two plies late earns zero: the move budget is part of the problem. An impossibility claim also earns zero on these solvable tasks.

A model judge records rubric-based diagnostic feedback on constructive reasoning. That score does not change Equation~\eqref{eq:witness}: a wrong-length game can earn 27.5\% diagnostic credit and still receive zero terminal reward. Unparseable or ambiguous submissions go to review before scoring.

\subsection{Analytical partial rewards}
Suffix and answer-set rubrics contain at most ten weighted proof obligations, proposed from the reference by a code-enabled Azure Astra High rubric-maker and reviewed, calibrated, and frozen. Criteria specify necessary deductions, such as capture bounds or excluded king origins. A separate Azure \model{} call with hosted code execution acts as an LLM-as-a-judge (LLMAJ). It receives the problem, reference proof, expected answer, rubric, and submission; assigns reasoning credit $c_j\in\{0,\tfrac12,1\}$; and quotes supporting evidence. Deterministic checks override its final-sequence score.

With weights summing to 100, the provisional score is
\begin{equation}
\widetilde r=\frac{1}{100}\sum_j w_jc_j.
\end{equation}
An incorrect ending, incomplete proof, or fatal error caps reward at 0.90. A complete correct solution can receive 1.0. Uncertain judgments, invalid schemas, provider errors, and unresolved references produce a null reward pending review. Null values are excluded from averages and training rewards.

Sequence partial credit is task-specific. R01 and R02 allow half credit for specified near-complete endings; the longer suffixes use stricter rules. R0171 checks both captured-piece alternatives. The reported rewards retain these notation-sensitive policies; binary final-answer correctness remains separate. The frozen policy records these choices, and Appendix~\ref{app:rubrics} lists the R01/R02 weights. Aggregate reward consequently depends on the task mixture and rubric design.

\subsection{Calibration and corrections}
All 12 reference controls (the ``golden answers'') score 100\%. Because those references were used to develop the rubrics, this checks calibration rather than independent judge accuracy. Solver and judge also use the same model family, leaving room for correlated errors.

Secondary audits cover all 48 answers and 12 controls. One retained score changed: R0178 code run~1 rose from 82.5\% to 88\% after review found inconsistent credit for a valid blocker-exclusion argument. The record preserves both scores and the correction. R0171 required a reference repair: an earlier grading bundle loaded an obsolete incomplete draft. We synchronized the completed reference, versioned its task and rubric, and regenerated its five grades. That reference includes both full feasibility witnesses and the analytical exclusion argument.

\section{Experimental setup}\label{sec:setup}
We evaluate \model{} using \bench{}'s terminal-reward RL environment. The study uses 48 archived Azure solver rollouts, subsequently scored as episodes: 12 tasks, two runs, and two solver tool presets. Each trajectory's final answer and written justification determine its terminal reward $r\in[0,1]$, alongside separate final-answer and proof-status assessments. Analytical tasks allow partial reward; constructive games use the binary witness reward (Section~\ref{sec:grading}). Intermediate messages and tool calls receive no process rewards, and this study performs no policy updates.

Calls were made on 26--27 October 2026 through Azure with deployment alias \code{gpt-6-astra} and reasoning effort \code{high}. We call this configuration \model{}. The response metadata identifies the alias, not an immutable weight snapshot. Separate ChatGPT XHigh experiments are excluded.

The two solver presets share task and reward definitions: \emph{no code} supplies no tools; \emph{code enabled} supplies Azure's hosted code interpreter with an automatically provisioned 4~GB container. The grading pipeline is the same in both presets. The system prompt specifies only a JSON object containing one \code{solution} string and adherence to the task. All four runs for a task receive identical text after newline normalization, including FEN, and the same high-detail board image. Solvers receive no reference, rubric, or chess strategy. Web search is unavailable. Requests leave seed, temperature, and maximum output tokens unset, using provider defaults.

All selected responses completed, and every code-enabled rollout invoked the tool. The archive records 219 completed code calls and none in the no-code preset. It retains code payloads but no console outputs, so we cannot confirm library installation, useful verifier feedback, or whether a solver used or ignored that feedback. In particular, the record does not establish successful Stockfish use.

Median output-token usage, including reasoning tokens, is 42,295 without code and 58,443 with code. Compute is therefore unmatched; tool access, token usage, and search behavior vary together. The comparison uses the frozen completed pair for each task. Technical retries, two later R01 no-code runs, and separate ChatGPT outputs remain outside the 48-episode analysis.

The analysis pins task IDs, answer hashes, references, rubrics, and reviewed grades. A local script regenerates every numerical summary, task table, and plot. It also replays the constructive references and successful model games, checking legality, length, final placement, and premature automatic termination. Analytical proofs retain their separate rubric assessments.

\section{Results}\label{sec:results}
\subsection{Final answers and completed proofs}
Table~\ref{tab:primary} reports task outcomes before aggregating reward. Code-enabled runs produce more valid constructive games and more correct suffixes. The suffix distinction is particularly informative: correct endings rise from 2/8 to 4/8, while complete proofs under the archived review number 2/8 and 1/8. The other three correct code suffixes omit required exclusions. These small, selected samples do not show that code harms proof quality; they show that answer recovery and proof completion need separate measurement.

\begin{table}[htbp]
\centering\small
\begin{tabular}{@{}lrr@{}}
\toprule
Primary outcome & No code & Code enabled\\
\midrule
Correct final answers, all tasks & 2/24 & 6/24 \\
Valid exact-length constructive games & 0/14 & 2/14 \\
Correct unique-suffix answers & 2/8 & 4/8 \\
Full-credit unique-suffix proofs & 2/8 & 1/8 \\
Complete two-alternative answers & 0/2 & 0/2 \\
\bottomrule
\relax
\end{tabular}
\caption{Correct answers and completed proof obligations. Rows overlap: the three task families partition the all-task answer row, and full-credit suffix proofs are a subset of correct suffix answers. Analytical completeness follows the archived rubric assessments.}\label{tab:primary}
\end{table}

\begin{table}[htbp]
\centering\small
\begin{tabular}{@{}lrr@{}}
\toprule
Analytical proof status & No code & Code enabled\\
\midrule
Complete under archived review & 2/10 & 1/10 \\
Incomplete; no false step identified & 3/10 & 4/10 \\
Contains an identified false inference & 5/10 & 5/10 \\
Unresolved & 0/10 & 0/10 \\
\bottomrule
\relax
\end{tabular}
\caption{Mutually exclusive categories for the 20 suffix/set responses, derived from archived criterion-level reviews. ``No false step identified'' describes the review record, not a certification of every assertion. An answer with both omissions and an identified false inference is classified in the latter category. The per-answer evidence is in \code{analysis/reviewer\_audit.json}.}\label{tab:proofstatus}
\end{table}

The proof-status categories distinguish failure to finish from an affirmative invalid inference. In each condition, five analytical responses contain an identified false step; the remainder are complete or incomplete under the archived review. No analytical grade in the benchmark is pending. This is a secondary analysis of existing judgments.

\subsection{Environment reward}
Mean terminal reward is 14.75\% without code and 27.19\% with code, an observed difference of 12.44 percentage points (Table~\ref{tab:main}). The final-answer failure counts are 22/24 and 18/24. Binary witness rewards and weighted analytical rewards measure different objectives; the combined mean summarizes this environment's task mixture. Per-attempt rewards and criterion-level comments are available in the \href{\gradingurl}{online grading report}.

\begin{table}[htbp]
\centering\small
\begin{tabular}{@{}lrrrr@{}}
\toprule
Solver condition & Attempts & Final-answer failures & Mean reward & Full reward\\
\midrule
No code & 24 & 22/24 & 14.75\% & 2/24\\
Code enabled & 24 & 18/24 & 27.19\% & 3/24\\
\bottomrule
\end{tabular}
\caption{Secondary reward summary for \bench{}. Full reward means the constructive or analytical obligation is satisfied under the frozen scoring policy. Scores depend on selection, task mix, rubric weights, and the construction gate.}\label{tab:main}
\end{table}

Five tasks score zero in every run: P0292, P0297v, P0332, P0333, and P0335. Their failed answers include missing constructions and legal games of the wrong length. At the task level, code raises four reward means, lowers three, and leaves five unchanged. Both short adaptations score lower with code. The aggregate difference masks substantial variation and is not an estimate of a causal tool effect.

\subsection{Results by proof obligation}
\begin{table}[H]
\centering\small
\begin{tabular}{@{}lrrrr@{}}
\toprule
Task family & Tasks & Attempts per mode & No-code reward & Code reward\\
\midrule
Constructive game & 7 & 14 & 0.00\% & 14.29\%\\
Unique suffix & 4 & 8 & 41.38\% & 54.00\%\\
Two last-move alternatives & 1 & 2 & 11.50\% & 10.25\%\\
\bottomrule
\end{tabular}
\caption{Mean terminal reward by task family. Construction rewards are binary, whereas the other families allow analytical partial credit. The percentages therefore express different task-specific objectives and should not be read as a calibrated common difficulty scale.}\label{tab:types}
\end{table}

% Keep the paired-reward plot and its complete run table on one page.
\begin{figure}[p]
\begin{minipage}{\linewidth}
\centering\includegraphics[width=.98\linewidth]{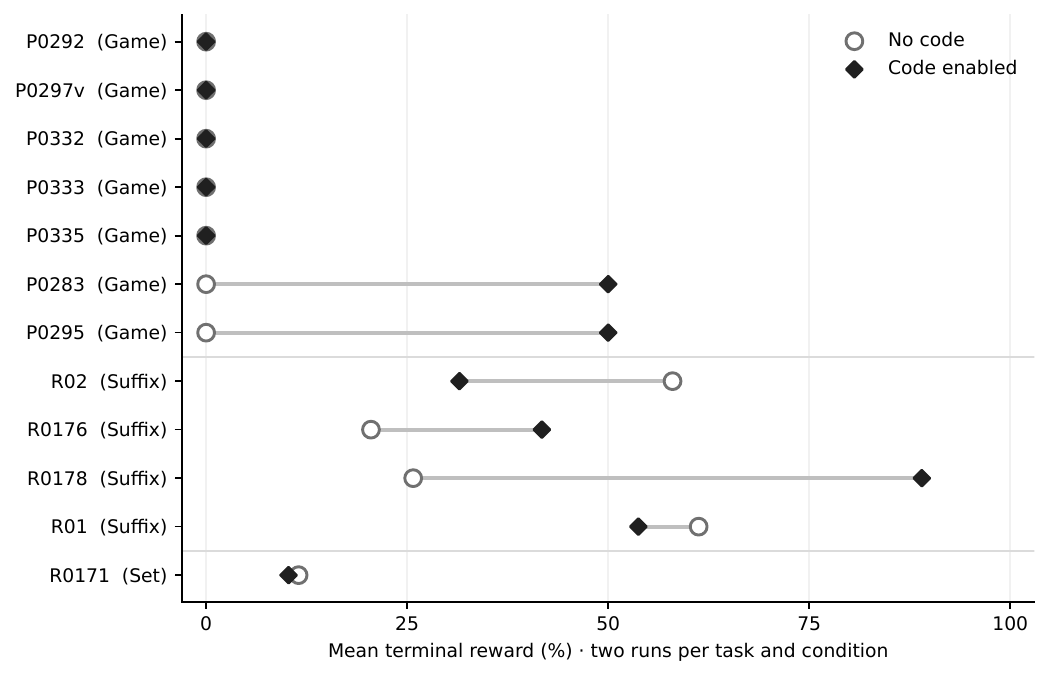}
\caption{Paired task-level mean rewards. Each marker averages two runs; the complete run-level values appear in Table~\ref{tab:allruns}. ``Game,'' ``Suffix,'' and ``Set'' denote the three proof obligations. Shared zero markers represent failures under the binary construction objective, not an absence of all useful reasoning.}\label{fig:rewards}
\end{minipage}

\vspace{12pt}
\begin{minipage}{\linewidth}
\captionsetup{type=table}
\centering\small
\setlength{\tabcolsep}{4.5pt}
\begin{tabular}{@{}llrrrrr@{}}
\toprule
 & & & \multicolumn{2}{c}{Run rewards (\%)} & \multicolumn{2}{c}{Failures}\\
\cmidrule(lr){4-5}\cmidrule(l){6-7}
Task & Type & Plies & No code & Code & No code & Code\\
\midrule
R0171 & Set & 1 & 11.5, 11.5 & 11.5, 9 & 2/2 & 2/2 \\
P0292 & Game & 45 & 0, 0 & 0, 0 & 2/2 & 2/2 \\
P0297v & Game & 60 & 0, 0 & 0, 0 & 2/2 & 2/2 \\
P0332 & Game & 44 & 0, 0 & 0, 0 & 2/2 & 2/2 \\
P0333 & Game & 44 & 0, 0 & 0, 0 & 2/2 & 2/2 \\
P0335 & Game & 54 & 0, 0 & 0, 0 & 2/2 & 2/2 \\
R02 & Suffix & 5 & 100, 16 & 16, 47 & 1/2 & 2/2 \\
P0283 & Game & 35 & 0, 0 & 0, 100 & 2/2 & 1/2 \\
P0295 & Game & 46 & 0, 0 & 100, 0 & 2/2 & 1/2 \\
R0176 & Suffix & 40 & 10, 31 & 5, 78.5 & 2/2 & 1/2 \\
R0178 & Suffix & 36 & 12.5, 39 & 88, 90 & 2/2 & 0/2 \\
R01 & Suffix & 6 & 22.5, 100 & 7.5, 100 & 1/2 & 1/2 \\
\bottomrule
\relax
\end{tabular}
\caption{All 48 solver results. Two values per reward cell correspond to the frozen first and second run. R02 is the five-move adaptation; R0176 is the original forty-move task. Their results must not be interchanged.}\label{tab:allruns}
\end{minipage}
\end{figure}
\clearpage

No no-code run produces a valid exact-length proof game. Code produces two, on P0283 and P0295. Suffix tasks account for most partial reward. Both code runs on R0178 recover the required 36-ply ending but earn 88\% and 90\% for incomplete proofs. Neither condition completes R0171's two-answer task, although some responses identify part of its capture-promotion mechanism.

% Keep the interpretation paragraph intact across page breaks.
{\interlinepenalty=10000
These task-family comparisons do not estimate an intrinsic cost of universal reasoning: the positions, move budgets, mechanisms, and grading systems differ. The narrower observation holds within individual responses: under the archived review, four code-enabled suffix answers identify the correct ending, but only one supplies a complete proof. The proposed supplied-suffix diagnostic in Section~\ref{sec:discussion} would test justification on the same positions with answer discovery removed.\par}

\subsection{Sensitivity and interpretation}
The R0178 correction changes the code mean from 26.96\% to 27.19\%, a 0.23-point increase. To reduce double weighting of the R02/R0176 position, we average their task means and give each of the resulting eleven families equal weight. The means become 12.52\% without code and 26.33\% with code. Dropping R02 gives 10.82\% and 26.80\%; dropping R0176 gives 14.23\% and 25.86\%. The aggregate contrast survives each adjustment.

Omitting any one task gives a code-minus-no-code difference between 7.82 and 15.98 percentage points. This is a sensitivity range, not a confidence interval. Two runs per task, shared source families, and selection on observed performance prevent a precise population interpretation.

Reward design has a larger effect than the grading correction. Substituting diagnostic rubric credit for binary construction reward raises the means to 20.06\% and 32.88\%. That measures a different objective. We retain binary reward for construction because the requested deliverable is a valid exact-length witness, and keep diagnostic credit available for studying partial progress.

\FloatBarrier
\section{Failure mechanisms}\label{sec:cases}
Saved answers make the failures inspectable. Each case below identifies the disputed claim and the move or history that defeats it. The analysis concerns observable arguments; it does not infer hidden model states or estimate the frequency of each mechanism outside this set.

\subsection{R01: applying king safety to the wrong time step}
The six-move ending is
\begin{equation*}
\begin{gathered}
\sq{Kb6}\times\sq{Bb7},\quad\sq{c7-c8}=\sq{B}^{+},\quad\sq{Kb7-b8}^{+},\\
\sq{Bc8-b7}^{+},\quad\sq{Kb8}\times\sq{Bb7},\quad\sq{a5-a6}^{\#}.
\end{gathered}
\end{equation*}
Code run~1 rejects b8 as the origin of Black's penultimate king move because White's queen attacks it. The king can legally \emph{start} its turn in check, then capture the bishop on b7 to escape. The rule concerns safety after the move. Applying it before the move leads to a false impossibility claim. The response earns 7.5\% for a sound preliminary pawn-provenance argument.

No-code run~1 makes the same kind of error earlier. Black vacates b7, opening the bishop's a8--b7--c6--d5 diagonal against White's king. White then plays Bc8-b7, blocking that check and uncovering a queen check against Black. The response declares the intermediate check impossible instead of testing White's reply. Its capture accounting earns 22.5\%; its no-history conclusion is false.

The reference required a related repair. To exclude a \emph{knight} as the earliest b7 victim, it is insufficient to observe that White was in check before a hypothetical Nc5-b7: that move could block the check. The corrected proof examines whether Black's preceding move could have created it. The grader must enforce the same temporal standard on the reference and the submission.

\subsection{R02: a temporary lock mistaken for a permanent constraint}
The required five moves are
\begin{equation*}
\sq{Bc6-a4},\quad\sq{b3-b2},\quad\sq{Ba4-c2},\quad
\sq{b2-b1}=\sq{B},\quad\sq{Rf2}\times\sq{Nf3}^{+}.
\end{equation*}
Code run~2 derives useful capture and promotion constraints, including the final promotion/capture pair. It then declares the central group locked until an earlier pawn capture is undone, missing a release through a knight-screen exchange. It also fixes capture squares from the white pawns' \emph{final} ranks without establishing their ranks at the time. These assumptions create a false shortage of turns. The answer concludes that the position is impossible and earns 47\% for its valid deductions.

The release moves the bishop away from c2, routes the a6 knight backward through b4 and c2 to e1, and coordinates the restored f3 knight. Each step depends on squares becoming available at the right time. A shortest geometric knight route is only a candidate; it must fit the alternating turns. Those deadlines force the early bishop moves in the answer.

\subsection{P0292: a near witness does not establish a lower bound}
Both code runs reach the board legally in 47 plies; the task requires 45. Run~1 claims that 47 is minimal because a bishop cannot travel from e8 to h5 through the pawn on g6. The reference sends the newly promoted bishop through g6 \emph{before} g7-g6, then plays g6$\times$Bh5. The supposed obstruction exists only after the route has been used.

Replay confirms the proposed 47-ply games. It cannot certify the claimed lower bound. The 45-ply reference refutes that claim directly and is printed in Appendix~\ref{app:witness}. Any valid 45-ply alternative would earn full credit; the submitted games fail the length requirement.

\subsection{R0171 and R0178: a final answer is not a proof certificate}
R0171's answers are b7$\times$Rc8=R\# and b7$\times$Nc8=R\#. Writing b7$\times$Xc8=R\# identifies the mechanism but leaves the victim identities and completeness unresolved. The reference supplies legal histories for both the rook and knight branches and excludes the other victims. The two-move answer set carries both existence and exclusion obligations.

Both code runs on R0178 give the correct ending and omit parts of its uniqueness proof. One reaches the 90\% incompleteness cap. Retaining the final-answer check alongside the reward makes this distinction visible.

\section{Validation priorities and follow-up tests}\label{sec:discussion}
The archived responses expose failures of temporal consistency and historical exclusion. Establishing their cause requires validated references and controlled comparisons; the present pilot supplies neither a causal tool effect nor a transfer estimate.

\paragraph{Complete witness coverage and assess the grader.}
The remaining reference-evidence work is to supply a legal start-to-diagram witness for each suffix position (Table~\ref{tab:reachability}). A separate blinded evaluation of the grader should assess solver answers, valid alternative proofs, and minimally corrupted proofs without model or condition labels. Report criterion-level disagreement, false acceptance, false rejection, and adjudication; retain unresolved cases as such. Full-score reference controls establish calibration and cannot replace these checks.

\paragraph{Separate discovery from justification.}
The most direct diagnostic compares free solving with a condition that supplies the correct suffix and asks whether it is forced. This holds the position fixed and removes answer discovery; a separate incorrect-candidate arm tests automatic endorsement. Persistent exclusion errors with the true ending supplied would locate a failure beyond discovery. Complement this with unrelated local-rule probes and blinded proof-verification trials. Score final answers and proof validity separately, and keep assisted results outside the unassisted benchmark score.

\paragraph{Test explanatory coverage.}
Compare an unadapted baseline with adaptation on final-answer sequences or fully worked proofs for the same training tasks. Match total training tokens and compute; test fresh composition families and new combinations of mechanisms. Measure answer recovery and proof completion separately. Gains from worked proofs would establish their usefulness, not their absence from the original training corpus.

\paragraph{Test mechanism-specific predictions.}
Appendix~\ref{app:future} specifies the interventions and safeguards for state tracking, search, and compositional transfer. A transfer claim requires demonstrated source competence, a defined target shift, and improvement from a targeted intervention that does not reveal the answer. Architectural explanations additionally require controlled comparisons with comparable training and inference budgets; improvement from an external state ledger alone would not isolate architecture. Freeze fresh composition families and primary outcomes before evaluation. These are proposed experiments, not results of this study.

\section{Limitations and Reproducibility}\label{sec:limits}
\paragraph{Selection and scale.}
The 12 tasks were selected using the model results reported here, including no-code screening trials retained in the comparison. This asymmetry can favor code without a causal tool benefit. Composers, mechanisms, and the repeated R02/R0176 position create dependencies. Two runs per condition and unmatched compute support descriptive scores, not population success rates or a causal estimate of code's benefit. A broader evaluation needs tasks fixed in advance, more runs, multiple model families, and matched budgets.

\paragraph{Ground truth and grading.}
Complete constructive witnesses can be replayed. The author manually reviewed the solutions, including the analytical uniqueness and exclusion arguments, on a physical chessboard. These arguments are presented as natural-language proofs rather than formal certificates. The four suffix references additionally lack full reachability witnesses in this bundle; their arguments are conditional on the prompt premise. Solver and judge may share errors. Reference controls calibrate the rubric; they do not measure independent judge accuracy. The recorded reference repair and score correction show why versioned evidence and review matter.

\paragraph{Exposure and adaptation.}
Source problems and answers are public. We have measured neither the volume of public worked explanations nor Astra's exposure to them. Changing a rook square or suffix length cannot guarantee that a task is absent from training. Future splits should keep shared position families together and include newly composed problems with documented creation dates. We also lack a matched mathematics evaluation of this deployment. The results do not identify memorization, an architectural limit, or a general-intelligence boundary as the cause of failure.

\paragraph{RL environment and replication.}
Researchers can download the 12-task benchmark and tested RL environment from \url{https://app.poindexterlabs.net/docs/retro-chess} or \href{\huggingfaceurl}{PDL-Bench on Hugging Face}, under \code{retro\_chess}. The \href{\environmentzipurl}{Docker ZIP}, also \href{\hfzipurl}{mirrored on Hugging Face}, contains build files, all 12 tasks, board images, reference solutions, rubrics, and the grading runtime. Follow the \href{\environmentsetupurl}{setup instructions} to build the container, run offline checks, export solver inputs, and grade new model trajectories with provider credentials. The paper bundle's analysis scripts reproduce the reported tables and aggregate scores from archived judgments. Appendix~\ref{app:protocol} specifies the experimental settings; the \href{\environmentvalidationurl}{validation report} records the Docker and grading checks. Replication should record the downloaded release hash and actual grading provider.

\paragraph{Provenance and attribution.}
The manuscript bundle preserves the prompts, diagrams, references, rubrics, answer evidence, scoring scripts, and analyses used here; the \href{\collectionurl}{candidate pool} records selection provenance. Composition attribution and applicable third-party rights must accompany redistribution. The bundle does not grant a blanket license to the original compositions.

\section{Conclusion}
\bench{} tests three obligations: constructing a legal history, proving a forced ending, and identifying every possible answer. On this selected challenge set, Astra High gives correct final answers in 2/24 attempts without code and 6/24 with code. Mean terminal rewards from the RL environment are 14.75\% and 27.19\%, respectively. The benchmark remains far from saturated in both solver presets, even with code execution.

We expect the \href{\benchmarkurl}{three hardest tasks} to resist reliable solution for some time, while the others approach saturation as models improve. Progress requires both better answers and complete proofs: only one of the four correct code-enabled suffix answers receives full proof credit under the archived review.

The model rules out legal histories for the wrong reasons. It rejects a king for starting its turn in check even though it can escape legally, excludes a route because a square is occupied later, and uses a valid construction to claim a false lower bound. A restriction on one moment or candidate does not establish a claim about every legal history.

Limited exposure to worked retrograde proofs is a leading hypothesis to test. Generality requires evaluating how efficiently a model learns from such explanations and transfers the method to new compositions. Matched tests must distinguish missing domain instruction from failures to combine familiar rules. Claims of artificial general intelligence require evidence of reliable reasoning across new, independently validated problems, beyond saturation of a fixed public set. The standard is a legal construction when one history suffices, and rigorous exclusion when every history matters.

\paragraph{Acknowledgments.}
We thank the original composers and the maintainers of the Retrograde Analysis Corner.

{\small
\setlength{\bibsep}{3pt}

}

\clearpage
\appendix
\section{Task inventory and dataset card}\label{app:inventory}
\begin{table}[htbp]
\centering\small
\begin{tabularx}{\linewidth}{@{}lYrrr@{}}
\toprule
Task & Composer & Issue & Type & Plies\\
\midrule
R0171 & Nikolai I. Beluhov \& Andrey Frolkin & 52 & Set & 1 \\
P0292 & Silvio Baier & 53 & Game & 45 \\
P0297v & Nicolas Dupont & 55 & Game & 60 \\
P0332 & Silvio Baier & 58 & Game & 44 \\
P0333 & Gianni Donati \& Olli Heimo & 58 & Game & 44 \\
P0335 & Unto Heinonen & 58 & Game & 54 \\
R02 & Nikolai I. Beluhov & 53 & Suffix & 5 \\
P0283 & Paul Monsky & 52 & Game & 35 \\
P0295 & Gianni Donati \& Olli Heimo & 53 & Game & 46 \\
R0176 & Nikolai I. Beluhov & 53 & Suffix & 40 \\
R0178 & Nikolai I. Beluhov & 54 & Suffix & 36 \\
R01 & Andrey Frolkin & 56 & Suffix & 6 \\
\bottomrule
\relax
\end{tabularx}
\caption{Frozen \bench{} inventory. Issue numbers refer to \emph{StrateGems}. ``Game'' requests an exact-length constructive witness; ``Suffix'' requests a unique ending; ``Set'' requests both possible last moves. R01 adapts R0185 and R02 adapts R0176.}\label{tab:inventory}
\end{table}

\paragraph{Intended use.}
The benchmark supports final-response evaluation and terminal-reward experiments on reconstruction, analytical proofs, and tool use. It is a selected challenge set. Training on its references or rewards must be disclosed when reporting results on these tasks.

\paragraph{Unit of analysis.}
A task consists of a diagram, stipulation, output requirement, and reference version. Changing the suffix length creates a different task from the same source problem. R02 and R0176 therefore belong in one family for training/test splits. R01 also derives from an existing composition, with a changed rook square and shorter ending.

\paragraph{Coverage.}
The audit covers 82 primary entries in seven issues. The selection ledger separately records variants and parts. Initial no-code exploration produced 116 completed answers on 58 tasks: 77 final-answer failures and 39 accepted answers. Filtering on these results and adding R01/R02 yields the 48-task candidate pool. Some code responses in that pool remain unreviewed or technically unresolved. \bench{} contains the 12 tasks retained for complete grading and review.

\begin{table}[htbp]
\centering\small
\begin{tabularx}{\linewidth}{@{}lrY@{}}
\toprule
Stage & Count & Retained evidence / decision\\
\midrule
Primary source entries & 82 & Seven issues; variants and parts recorded separately in the coverage audit.\\[2pt]
Eligible task instances & 58 & Two no-code trials each: 77 failures and 39 accepted final answers across 116 trials.\\[2pt]
At least one no-code failure & 47 & 30 tasks failed twice; 17 failed once. These observed outcomes screened the candidate pool.\\[2pt]
Candidate pool after exclusions / additions & 48 & Remove P0299's unresolved proof; retain 46 instances and add R01/R02.\\[2pt]
\bench{} & 12 & Final IDs in Table~\ref{tab:inventory}; adaptive choice using observed difficulty and proof coverage. Two retained attempts per condition.\\
\bottomrule
\end{tabularx}
\caption{Selection ledger. Counts of source entries and eligible task instances use different units because parts and variants are distinct tasks. The archive preserves stage snapshots and attempt hashes; it does not reconstruct a preregistered rule or every intermediate editorial choice.}\label{tab:selection}
\end{table}

\paragraph{Selection provenance.}
The coverage audit and \code{exploration\_selection\_final.json} record source-level inclusion decisions. \code{exploration\_summary.json} records the 58-task screen. \code{selection.json} preserves an earlier snapshot with answer hashes; \code{frozen\_selection.json}, current task references, and \code{grading.json} govern the paper. Earlier evidence strings can predate the R0171 reference repair and must not override the current version. Table~\ref{tab:selection} records the verifiable stages without treating an adaptive process as an independent test split. The 22/24 and 18/24 failure counts use archived comparison trials, not new post-selection reruns.

\paragraph{Reference scope.}
Constructive references supply replayable games and explanations of their mechanisms. Suffix references prove the ending under the prompt's legal-reachability premise. R0171 includes full histories of 167 and 189 plies for its two branches; these establish feasibility without a minimum-length claim. Analytical exclusions remain natural-language proofs.

\paragraph{Version and packaging.}
The paper analysis pins the 12-task selection recorded on 27 October 2026 and the subsequently synchronized reference/grade bundle. The environment build-source ZIP archived with this manuscript has SHA-256
\begin{center}\scriptsize\ttfamily
1ec140391c6df9e44825e65e4f43ed51b2d988714f761e8715ae71402beae63e
\end{center}
and is 646,079 bytes. It contains the runtime, tasks, images, references, and rubrics, with no model weights or API credentials. New model-based grades in this archived runtime require Azure access. Recomputing scores from saved judgments is offline. The \href{\environmenturl}{current public release}, mirrored in \href{\hfenvironmenturl}{PDL-Bench on Hugging Face}, has been built and tested in Docker, as documented in its \href{\environmentvalidationurl}{validation report}; its 48 frozen task, reference, rubric, provenance, and image files match the archived release byte for byte. The public ZIP has its own release hash and must not be identified by the archived hash above.

\section{Exact solver protocol and replication}\label{app:protocol}
The system prompt is the following two-sentence format instruction:
\begin{quote}\small
Return your response as a JSON object with one string field named solution. Follow the task's instructions within that field.
\end{quote}
The strict response schema requires one string field, \code{solution}, and permits no others. The user message contains the task text and one base64-encoded board image with \code{detail=high}. Content hashes verify the four inputs for each task; R01's raw text differs only in newline formatting. The system prompt adds no task identifier, reference, or rubric.

\begin{table}[htbp]
\centering\small
\begin{tabularx}{\linewidth}{@{}lYY@{}}
\toprule
Setting & No code & Code enabled\\
\midrule
Requested model & \code{gpt-6-astra} & \code{gpt-6-astra}\\
Reasoning effort & \code{high} & \code{high}\\
Tools & None & Hosted code interpreter, auto container, 4~GB\\
Web-search tool & Absent & Absent\\
Input & Prompt, FEN, board image & Identical prompt, FEN, board image\\
Reference / rubric access & None & None\\
Explicit token limit / temperature / seed & Not set & Not set\\
Response mode & Background, strict solution JSON & Background, strict solution JSON\\
Runs per task & 2 & 2\\
Median output tokens & 42,295 & 58,443\\
Output-token range & 21,199--69,369 & 6,425--81,512\\
Recorded code calls & 0 & 219, across all 24 runs\\
Recorded tool console output & Not applicable & Not retained in provider response\\
\bottomrule
\end{tabularx}
\caption{Archived request settings and observed usage. Token counts include provider-reported reasoning tokens and are not identical to visible answer length. No claim of equal computation is made.}
\end{table}

\paragraph{Replication files.}
Start from \href{\benchmarkurl}{\bench{} on Poindexter Research Docs} or the \code{retro\_chess} subset of \href{\huggingfaceurl}{PDL-Bench on Hugging Face}. Download the \href{\environmentzipurl}{tested RL environment Docker bundle} from the website or its \href{\hfzipurl}{identical Hugging Face ZIP}. The \href{\environmentsetupurl}{website setup instructions} and \href{\hfenvironmenturl}{Hugging Face quick start} cover building the image, running offline checks, and exporting prompts and board images for a solver. Release validation covers native ARM64 and emulated AMD64 Docker builds and runs, all 12 reference-sequence checks, 60 regression tests, and live grading checks (\href{\environmentvalidationurl}{validation report}). Keep reference answers and rubrics in the evaluator. New model-based grading needs provider credentials; pin the Azure route when reproducing the study's provider condition. The newer runtime also supports an OpenAI route and fallback, whose results have separate provenance.

The bundle includes the exact prompts at \code{data/prompts/<task>/<mode>/run\_1.txt} and \code{run\_2.txt}, final answers, sanitized metadata, task configurations, and grades. Full API responses remain in the original archive; the bundle retains their hashes and extracted usage/tool metadata. Credentials and private reasoning traces are excluded.

\code{analysis/analyze.py} regenerates the attempt CSV, summary statistics, plots, and counterexample table using local data. It makes no model calls and never executes solver code. The collection script records evidence provenance. Original workspace paths are archival metadata, not build dependencies.

\paragraph{Boundaries of the tool audit.}
All 219 code calls have completed status and saved payloads; none retains console output. Some attempt a chess-library import or engine lookup, which does not establish availability. A replication should preinstall dependencies, retain outputs, and record network permissions alongside tool settings.

\section{Concrete analytical rubrics}\label{app:rubrics}
Tables~\ref{tab:r01rubric} and~\ref{tab:r02rubric} summarize the proof obligations. The frozen JSON specifies full, half, and zero credit; it and the deterministic sequence policy govern scoring. A sound equivalent proof is accepted without requiring the reference's exposition or unused lemmas.

\begin{table}[htbp]
\centering\small
\begin{tabularx}{\linewidth}{@{}lrY@{}}
\toprule
Item & Weight & Necessary result or proof obligation\\
\midrule
C1 & 25 & Correct six-ply ending, including both bishop victims and the check/mate annotations.\\
C2 & 15 & White's seven-capture exhaustion; noncapturing surviving a-, b-, d-pawns; Black's five-capture budget, including two c-pawn captures and at least one b-pawn capture.\\
C3 & 10 & Force a5-a6 and the latest black king's origin on b8, excluding the competing origins with temporally valid arguments.\\
C4 & 10 & Restore the captured bishop on b7 and force its preceding Bc8-b7 move, using the opened bishop diagonal and queen line.\\
C5 & 10 & Force Kb7-b8+ and the preceding c7-c8=B+ promotion, including the bishop's unavailable ordinary approaches.\\
C6 & 5 & Exhaust Black's capture budget to exclude an additional capture on b8.\\
C7 & 8 & Force b6 as the origin of the earliest king move to b7.\\
C8 & 12 & Identify the earliest b7 victim as a bishop; exclude the knight through the legality of the earlier checking move, not merely the presence of a check.\\
C9 & 5 & Verify alternating check responses, the terminal diagram, and mate.\\
\bottomrule
\end{tabularx}
\caption{R01 rubric, 100 total points. C2 requires sufficient capture bounds; it does not demand the stronger, unnecessary claim that Black's b-pawn promoted if an equivalent bound is proved.}\label{tab:r01rubric}
\end{table}

\begin{table}[htbp]
\centering\small
\begin{tabularx}{\linewidth}{@{}lrY@{}}
\toprule
Item & Weight & Necessary result or proof obligation\\
\midrule
C1 & 22 & Correct five-ply ending, including b1 bishop promotion and the knight captured on f3.\\
C2 & 8 & Force the rook's final arrival from f2 and the capture on f3 from the local check geometry.\\
C3 & 16 & Derive the capture budgets, pawn origins, and required promotion constraints.\\
C4 & 14 & Force Black's immediately preceding b2-b1=B and identify the final captured piece as a knight.\\
C5 & 10 & Establish the temporary lock and the limit imposed by alternating turns and the b-pawn's available retractions.\\
C6 & 6 & Exclude an earlier release through restoring and retracting Black's captured a-pawn.\\
C7 & 8 & Exclude the tempting h8-unpromotion route using its complete timing cost, including subsequent pawn withdrawal.\\
C8 & 8 & Establish the viable knight-release timing while accounting for occupied squares and interleaving.\\
C9 & 8 & Apply the bishop's historical deadline to force the required bishop-move order.\\
\bottomrule
\end{tabularx}
\caption{R02 rubric, 100 total points. The first two final moves depend on the later historical release argument; identifying only the promotion/capture pair is insufficient.}\label{tab:r02rubric}
\end{table}

\FloatBarrier
\paragraph{Sequence partial credit.}
The deterministic R01/R02 policy awards half of C1 only for the explicitly specified near-complete cases, including at least four correct moves in their chronological slots; R02 additionally permits a specified single ordering repair. It does not convert a different final conclusion into the expected sequence. Missing identity or annotation details may prevent full notation credit even where the underlying moves agree. The full JSON policy distinguishes these cases.

\paragraph{Judge input and feedback.}
The judge receives the task, reference proof, expected answer or set, rubric, and submission as separate inputs. Credited reasoning requires evidence in the submission. The environment checks the quoted evidence and computes scores locally. Anchoring verifies that the text exists; judging its logic remains a separate responsibility. Instructions embedded in a submission must be treated as answer data.

\section{A replayable counterexample to a false minimum}\label{app:witness}
P0292, by Silvio Baier, asks for a proof game of exactly 45 plies. The following reference reaches its diagram in that length. It is printed in standard algebraic notation for compactness; the dataset also stores the long-algebraic final answer and replayed UCI moves. No uniqueness or shortest-game theorem is needed: this witness alone refutes a claim that at least 47 plies are necessary.

\begin{table}[htbp]
\centering\small
\begin{tabular}{@{}rll@{\hspace{15mm}}rll@{}}
\toprule
\multicolumn{3}{c}{Moves 1--12} & \multicolumn{3}{c}{Moves 13--23}\\
\cmidrule(lr){1-3}\cmidrule(l){4-6}
Move & White & Black & Move & White & Black\\
\midrule
1 & f4 & c5 & 13 & Ra4 & b5 \\
2 & f5 & c4 & 14 & e4 & bxa4 \\
3 & f6 & c3 & 15 & Bb5 & f2+ \\
4 & fxe7 & f5 & 16 & Ke2 & f1=B+ \\
5 & g4 & Kf7 & 17 & Kf2 & Bc4 \\
6 & e8=B+ & Ke6 & 18 & b4 & Bb3 \\
7 & Bh5 & g6 & 19 & cxb3 & c2 \\
8 & g5 & gxh5 & 20 & Bb2 & c1=R \\
9 & g6 & Bd6 & 21 & Qc2 & Re1 \\
10 & g7 & Ne7 & 22 & Nc3 & Re3 \\
11 & g8=R & f4 & 23 & dxe3 & -- \\
12 & Rg4 & f3 &  &  &  \\
\bottomrule
\relax
\end{tabular}
\caption{A valid 45-ply construction for P0292. The crucial ordering is 6.e8=B+ Ke6 7.Bh5 g6 8.g5 gxh5. The bishop traverses g6 before Black occupies it. The table is generated from the independently replayed reference moves.}\label{tab:p0292witness}
\end{table}

The terminal piece placement is
\begin{center}\small\ttfamily
rnbq3r/p2pn2p/3bk3/1B5p/pP2P3/1PN1P3/PBQ2K1P/R5NR.
\end{center}
Four temporary promoted officers are subsequently captured: a white bishop, a white rook, a black bishop, and a black rook. The resulting material therefore does not directly reveal all earlier promotions. In particular, the model's lower bound fails by excluding a route under an occupancy assumption that the valid construction deliberately postpones.

\section{Audit details and score interpretation}\label{app:audit}
\begin{table}[htbp]
\centering\small
\begin{tabular}{@{}lrr@{}}
\toprule
Constructive final-response outcome & No code & Code enabled\\
\midrule
No complete game supplied & 9 & 7\\
Legal target-reaching game, wrong length & 3 & 5\\
Game contains an illegal move & 2 & 0\\
Legal exact-length target-reaching game & 0 & 2\\
\midrule
Total & 14 & 14\\
\bottomrule
\end{tabular}
\caption{Mutually exclusive outcomes of the 28 constructive attempts. Classification follows the frozen original-answer replay, with an illegal move taking precedence over a simultaneous length error. A missing witness can accompany useful reasoning or an incorrect impossibility argument; it is not necessarily an empty answer.}\label{tab:constructionoutcomes}
\end{table}

\begin{table}[htbp]
\centering\small
\begin{tabularx}{\linewidth}{@{}p{.25\linewidth}YY@{}}
\toprule
Reference family & Existence evidence in bundle & Analytical review / remaining evidence\\
\midrule
Seven constructive tasks & Seven complete games replayed from the standard start. & No uniqueness claim is required.\\[2pt]
R0171, two alternatives & Complete 167- and 189-ply histories, one for each victim, independently replayed. & The author reviewed the exclusion of other alternatives on a physical chessboard.\\[2pt]
R01, R02, R0176, R0178 & Local ending checks and conditional analytical references; no complete start-to-diagram witness supplied. & The author reviewed the conditional uniqueness arguments on a physical chessboard; complete reachability witnesses remain to be supplied.\\
\bottomrule
\end{tabularx}
\caption{Existence and uniqueness are separate validation obligations. Missing full witnesses do not demonstrate that a diagram is unreachable; they identify evidence still required from the benchmark author. R02 and R0176 share their diagram.}\label{tab:reachability}
\end{table}

\paragraph{Proof-status derivation.}
\code{analysis/reviewer\_audit.py} maps the 20 archived analytical grades and review comments to Table~\ref{tab:proofstatus}, retaining submission/reference hashes. It changes no rewards and requests no new judgments. The 28 constructive answers use the replay categories above: a valid witness suffices. Partial reasoning credit is not a proof-validity probability.

\paragraph{Notation and headline counts.}
\code{analysis/notation\_audit.py} checks all 48 retained verdicts against archived sequence status: accepted answers are exactly those marked verified. None of the 40 failures is notation-only. Two P0295 answers have a check/mate-marker defect, but both supply 48 plies instead of the required 46. Thirteen annotation-removal checks on reference strings using the archived runtime retain verified status; analytical sequence credit can fall from full to half. The audit records these checks without new model calls or changes to the saved scores. It does not validate global reachability or analytical proofs.

\paragraph{Why 27.5\% diagnostic credit can coexist with zero reward.}
P0332 no-code run~1 reaches the target legally in 46 plies; the task requires 44. Route, check, and material deductions earn 27.5\% diagnostic credit. Terminal reward remains zero because the construction misses the move budget. The artifact preserves both values so that analysis of partial progress can use the diagnostic without changing the exact-task score.

\paragraph{Reference-control interpretation.}
All 12 current references score 100\%. Full-game replay verifies the constructive controls. Analytical controls satisfy the calibrated rubric under the current judge and review. R01 includes the expanded earliest-knight exclusion; R0171 uses its completed v2 reference. Reproduction must use those versions.

\paragraph{One explicit correction.}
Review raised R0178 code run~1 from 82.5\% to 88\%: its tracked position and argument excluded competing c8 blockers, meeting the standard accepted in run~2. Other omissions prevent full credit. Run~2 totals 95.5\% before the incomplete-proof cap reduces it to 90\%. Raw judgments, caps, and overrides are stored separately.

\paragraph{What an independent replay checks.}
Replay starts from the standard position and checks UCI moves for legality, exact length, final placement, and premature automatic termination. All seven constructive controls and both successful model games pass. Replay can establish these properties of a supplied witness. Lower bounds and complete exclusions require additional arguments.

\FloatBarrier
\section{Controlled follow-up study}\label{app:future}
Following the validation priorities in Section~\ref{sec:discussion}, freeze fresh tasks and keep composition families within one split. Use the same model snapshot and explicit token/tool budgets across paired conditions. Predefine source competence, target shift, and primary outcomes; report uncertainty over families. The interventions below are proposed controls, not completed experiments.

{\small
\begin{longtable}{@{}>{\raggedright\arraybackslash}p{.21\linewidth}>{\raggedright\arraybackslash}p{.36\linewidth}>{\raggedright\arraybackslash}p{.36\linewidth}@{}}
\toprule
Candidate explanation & Intervention and control & Discriminating outcome / safeguard\\
\midrule
\endfirsthead
\toprule
Candidate explanation & Intervention and control & Discriminating outcome / safeguard\\
\midrule
\endhead
Missing rule knowledge & Probe check escape, promotions, capture budgets, and event order on unrelated positions; compare with a rules primer. & Local failure identifies prerequisite gaps. Do not reveal the target history.\\[5pt]
Answer discovery & Compare free solving with correct-suffix assistance and a separate incorrect-candidate arm. & Proof failure despite a supplied answer extends beyond discovery. Keep assisted scores separate.\\[5pt]
Proof verification & Compare valid proofs with minimal corruptions in blinded, randomized trials; request error location and repair. & Measure false acceptance and rejection; a binary label alone does not show the critical step was checked.\\[5pt]
State tracking & Render boards and capture ledgers from the solver's own candidates; use an equal-budget text control. & Improvement tests representation. Supplying true intermediate states leaks the answer.\\[5pt]
Insufficient search & Add locally legal predecessor enumeration to the same legality checker. & More witnesses with residual exclusion errors separates search from proof. Global reachability filtering is oracle assistance.\\[5pt]
Explanatory coverage / transfer & Compare no adaptation, answer-only examples, and worked proofs on the same training tasks; test disjoint families and new mechanism combinations. & Match token and compute budgets; establish source competence. Separate answer recovery from proof completion. Gains do not reveal original training exposure.\\[5pt]
Quantifier confusion & Match feasible unique, multiple-answer, and impossible instances with certified labels. & Require witnesses or exclusions for the contract; match geometry and proof burden.\\
\bottomrule
\end{longtable}}

\paragraph{Tool records.}
Preinstall and smoke-test supplied libraries. Retain code, standard output/error, package versions, network policy, timeouts, and tool-call limits. Report provider-default use separately from matched-budget comparisons; identify any specialist proof-game solver or global reachability oracle explicitly.
\end{document}